%% file: root.tex
\documentclass[letterpaper, 10 pt, conference]{ieeeconf}  

\IEEEoverridecommandlockouts                              

\usepackage[inkscapeformat=png]{svg}

\usepackage{commenting}
\usepackage{microtype}

\usepackage{tabularx}

\usepackage{xcolor}
\definecolor{mypurple}{RGB}{150, 115, 166}
\definecolor{myorange}{RGB}{255, 199, 135}
\definecolor{myyellow}{RGB}{255, 223, 94}
\definecolor{myred}{RGB}{184, 84, 80}
\definecolor{myblue}{RGB}{108, 142, 191}
\definecolor{mygreen}{RGB}{130, 179, 102}

\usepackage{multirow}
\usepackage{hhline}

\usepackage{wrapfig}
\usepackage[most]{tcolorbox}
\definecolor{cocoabrown}{rgb}{0.82, 0.41, 0.12}
\usepackage{tikz}
\usepackage{changepage}
\definecolor{ao(english)}{rgb}{0.0, 0.5, 0.0}
\definecolor{burntsienna}{rgb}{0.91, 0.45, 0.32}

\usepackage{url}

\usepackage{subcaption}

\usepackage{hyperref}

\title{\LARGE \bf
  Chat with the Environment: Interactive Multimodal Perception Using Large Language Models
}

\author{Xufeng Zhao$^{\ast}$, Mengdi Li, Cornelius Weber,  Muhammad Burhan Hafez, and Stefan Wermter
\thanks{This research was funded by the German Research Foundation (DFG) in the project Crossmodal Learning (TRR-169) and the China Scholarship Council (CSC).}
\thanks{The authors are with the Knowledge Technology Group, Department of Informatics, Universität Hamburg, 22527 Hamburg, Germany. E-mail: $\{$xufeng.zhao, cornelius.weber, burhan.hafez, stefan.wermter$\}$@uni-hamburg.de, mengdi.li@studium.uni-hamburg.de.}
\thanks{$^{\ast}$Corresponding author.}}

\begin{document}

\maketitle
\thispagestyle{empty}
\pagestyle{empty}

\input{abstract}

\input{introduction}
\input{relatedwork}
\input{methodology}
\input{experiments}
\input{limitation}
\input{conclusion}

\bibliographystyle{plain} 
\bibliography{bib-extracted.bib,Matcha.bib}

\addtolength{\textheight}{-12cm}   






\end{document}

%% file: abstract.tex
\begin{abstract}

Programming robot behavior in a complex world faces challenges on multiple levels, from dextrous low-level skills to high-level planning and reasoning.
Recent pre-trained Large Language Models (LLMs) have shown remarkable reasoning ability in few-shot robotic planning. 
However, it remains challenging to ground LLMs in multimodal sensory input and continuous action output, while enabling a robot to interact with its environment and acquire novel information as its policies unfold.
We develop a robot interaction scenario with a partially observable state, which necessitates a robot to decide on a range of epistemic actions in order to sample sensory information among multiple modalities, before being able to execute the task correctly.
\textit{Matcha} (\textit{M}ultimod\textit{a}l environmen\textit{t} \textit{cha}tting) \textit{agent}, an interactive perception framework, is therefore proposed with an LLM as its backbone, whose ability is exploited to instruct epistemic actions and to reason over the resulting multimodal sensations (vision, sound, haptics, proprioception), as well as to plan an entire task execution based on the interactively acquired information.
Our study demonstrates that LLMs can provide high-level planning and reasoning skills and control interactive robot behavior in a multimodal environment, while multimodal modules with the context of the environmental state help ground the LLMs and extend their processing ability.
The project website can be found at \url{https://matcha-agent.github.io}.
\end{abstract}

%% file: introduction.tex
\section{Introduction}

\textit{How do humans perceive the surroundings to uncover latent properties?}

Suppose you are presented with an uncommon object in a strange shape and of unknown material,
you may explore its properties in both passive and active ways, if possible, e.g.\ by observing the geometry, touching and even knocking on the surface in order to deduce its exact functionalities from the feedback.
Unnecessary explorations, which could be essential for other scenarios such as smelling, will not be performed in this context unless something counterintuitive happens.
We humans naturally perform these \textbf{multimodal observations and examinations} in daily life through \textbf{common sense and established knowledge}, and over time we adapt with the accumulation of experience \cite{Barnard06CrossModal}.

This is also now possible for a robot well-equipped with multiple sensors and Large Language Models (LLMs).
An environment may be filled with rich content, and the robot can be overwhelmed with diversified sensory stimuli.
An intelligent robot should 1) wisely choose stimuli to attend to, avoiding eagerly being bogged down into details, and 2) respond accordingly to the resulting sensations in the context of a specific task.
\begin{figure}[t!] 
    \centering
    \includegraphics[width=\linewidth]{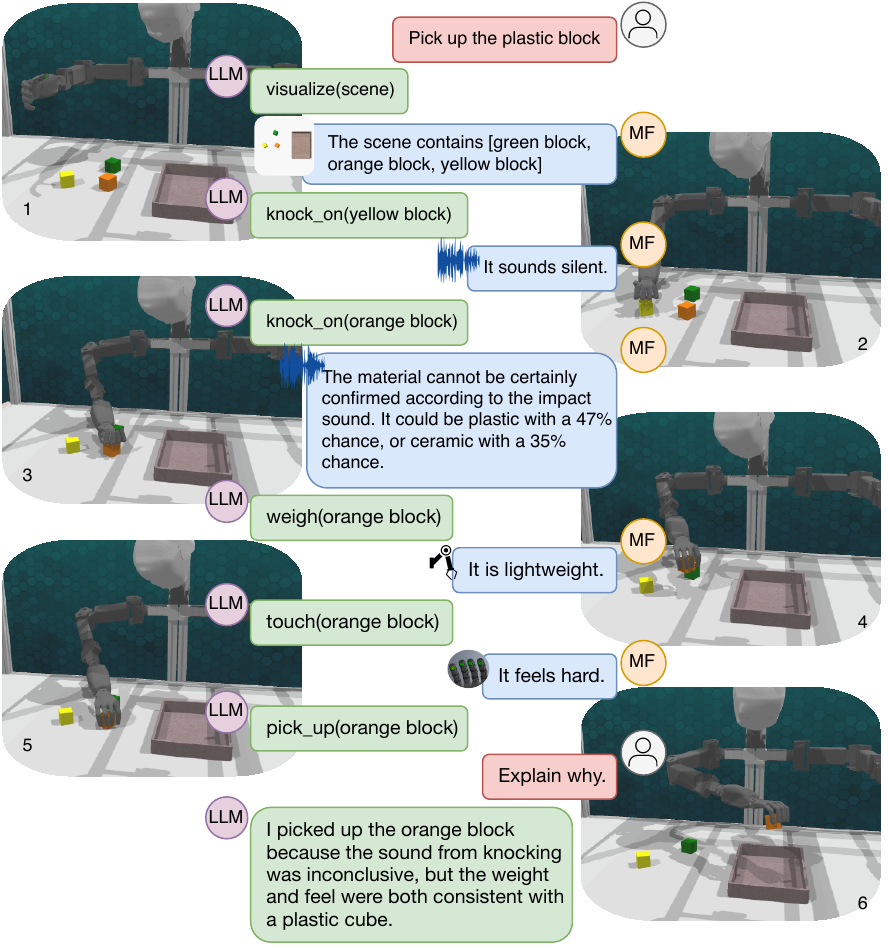}
    \caption{Given instruction from a human, the robot recurrently ``chats" with the environment to obtain sufficient information for achieving the task. An LLM generates action commands to interactively perceive the environment and, in response, the environment provides multimodal feedback (MF) through multimodal perception modules. }
    \label{fig:exp}
\end{figure}

\subsection{Interactive Multimodal Perceptions}

Like humans, robots can perceive the environment in either a passive or an interactive way \cite{Kroemer21ReviewRobot}.
\textit{Passive perception} refers to ways such as visual or auditory monitoring, and it allows robots to quickly acquire information without intervening with the surroundings. However, the passive manner has its limits, among which the most outstanding one is its impotency when facing \textit{epistemic uncertainty} \cite{Celemin23KnowledgeandAmbiguityaware}, the uncertainty because of lacking knowledge.

Epistemic uncertainty inevitably arises from diverse sources, e.g.\ from the ambiguity in human instructions, from low-resolution sensing (e.g. reduced image size for convolution), or from insufficient modalities.
Many of them can only be reduced with \textit{interactive perception}, in which a robot actively interrogates the environment to increase accuracy and even uncover latent information.
For example, when being asked to deliver a \textit{steel} screw instead of one with a similar color \& shape but made of \textit{aluminum}, an assistant robot may need to locate possible candidates with \textit{passive} vision and further, \textit{interactively}, resort to a weighing or a magnetic module for confirmation.

Despite the promising advantages, interactive perception is less common than the passive manner because it entails increased complexity \cite{Li23InternallyRewarded}.
Efforts are needed to design a mediating system to handle various sensory data and to adapt to changes in the conditions of both the robot and the environment, such as a new robotic modular being available or the involvement of novel objects.

\subsection{Chatting with the Environment}
LLMs have been showing incredible potential in areas besides robotics \cite{Ahn22CanNot, Cui23NoRight, Lynch22InteractiveLanguage, Mialon23AugmentedLanguage}.
Human knowledge that resides in LLMs can help a robot abstract and select only suitable features, e.g. relevant to the region of interest or informative modalities, to simplify the learning process.
Moreover, in terms of generalizability, the knowledge of LLMs allows a behavioral agent to adapt efficiently to novel concepts and environmental structures.
For instance, when being asked to \textit{use one adjective for each to describe how a sponge and a brick feel}, ChatGPT\footnote{\url{https://openai.com/blog/chatgpt/}} will respond with ``soft'' and ``hard'' respectively. This is helpful for a robot with a haptics sensing module to distinguish between these two novel, never-seen objects.

LLMs are usually generative models that predict tokens to come, but with certain designs, e.g.\ conversational prompts, LLMs are capable of generating chat-fashion texts. This allows their integration with a robot to not only plan with respect to a robot's built-in ability \cite{Zeng23SocraticModels, Ahn22CanNot}  but also respond according to environmental feedback.

However, they cannot directly process application-specified raw multimodal data. We 
resort to modular perceptions for each modality that are separately trained before being plugged into the LLM backbone. Each module semantically translates the resulting multimodal sensations into natural language that can be understood by LLMs and processed in a unified manner.

Our contributions are threefold. Firstly, we establish a manipulation scenario with multimodal sensory data and language descriptions. 
Secondly, we propose \textbf{Matcha}\footnote{By the name of a type of East Asian green tea.
To fully appreciate matcha, one must engage multiple senses to perceive its appearance, aroma, taste, texture, and other sensory nuances.
} (\textbf{M}ultimod\textbf{a}l environmen\textbf{t} \textbf{cha}tting) \textbf{agent}, where an LLM is prompted to work in a chatting fashion, thus having continuous access to environmental feedback for contextual reasoning and planning.
Finally, we show that LLMs can be utilized to perform interactive multimodal perception and behavior explanation. Accordingly, an interactive robot can make reasonable and robust decisions by resorting to LLMs to examine objects and clarify their properties that are essential to completing the task (see Fig.~\ref{fig:exp}). 

%% file: relatedwork.tex
\section{Related Work}

\textbf{Multimodal Learning and Robotic Information Gathering.}
Research in multimodality in robotics nowadays attracts growing attention \cite{Akkus23MultimodalDeep} because of its success in, for example, audio-visual learning \cite{Zhao22ImpactMakes, Wei22LearningAudiovisual, Zhu21DeepAudiovisual} and language-visual learning \cite{Shridhar22CLIPortWhat, Shridhar22PerceiveractorMultitask}. 
It is beneficial and sometimes essential for a robot to learn from multimodality because one modality could carry some distinct information, e.g.\ tones in speech, that cannot be deduced from another. \cite{Lee22SoundguidedSemantic}.

Capable robots require managing one or several sensors to maximize the information needed for disambiguation \cite{Barnard06CrossModal} regarding a specific goal. This problem is known as \textit{active information acquisition} \cite{Atanasov15ActiveInformation, Wakulicz21ActiveInformation} or, particularly in robotics, \textit{robotic information gathering} \cite{Rankin21RoboticInformation}, where robots have to properly select perceiving actions to reduce ambiguity or uncertainty.
Besides handcrafted rules, some information advantage measures, e.g.\ entropy or information gain, are usually employed to maximize \cite{Atanasov15ActiveInformation}.
However, the combination of multimodal data is usually challenging. There are studies on fusing multimodal data according to their uncertainties, but this may face numerical instability and is difficult to transfer from one application to another \cite{Wang22UncertaintyawareMultimodal}.
Instead of directly fusing the multisensory data in a numerical space, we propose to use multimodal modules to translate them into natural language expressions that an LLM can easily digest.

\textbf{Large Language Models in Robotic Planning}.
Very recent works use LLMs to decompose high-level instructions into actionable low-level commands for zero-shot planning. They use LLMs as a planner to autoregressively select actions that are appropriate with respect to the instruction according to application-based prompts \cite{Zeng23SocraticModels}, the semantic similarity between mapped pairs \cite{Huang22LanguageModels}, or the contextual language score grounded on realistic robot affordances \cite{Ahn22CanNot}.
Other approaches ground LLM knowledge in human interaction \cite{Cui23NoRight} or many other various fields where domain knowledge is distinct and modular frameworks can be composed via language as the intermediate representation \cite{Patki19InferringCompact, Mialon23AugmentedLanguage, Zeng23SocraticModels}.

However, these works design a robot to form a planning strategy with \textit{built-in knowledge}, rather than \textit{interact} with the surroundings and make decisions based on \textit{actively collected information} from the environment. There is no feedback loop for their LLMs to perceive the environmental cues, such that only ``blind" decisions are made in the robotic unrolling process. 
In contrast, our interactive architecture allows LLMs to access the environment state from multiple modalities for adaptive planning.

%% file: methodology.tex
\section{Methodology}

\subsection{Architecture}
We propose \textbf{Matcha} (\textbf{m}ultimod\textbf{a}l environmen\textbf{t} \textbf{cha}tting)  \textbf{agent} which is able to interactively perceive (``chat" with) the environment through multimodal perception when the information from passive visual perception is insufficient for completing an instructed task.
The epistemic actions are executed autoregressively until the agent is confident enough about the information sufficiency in that situation. 

\begin{figure}[thpb]
    \centering
    \includegraphics[width=\linewidth]{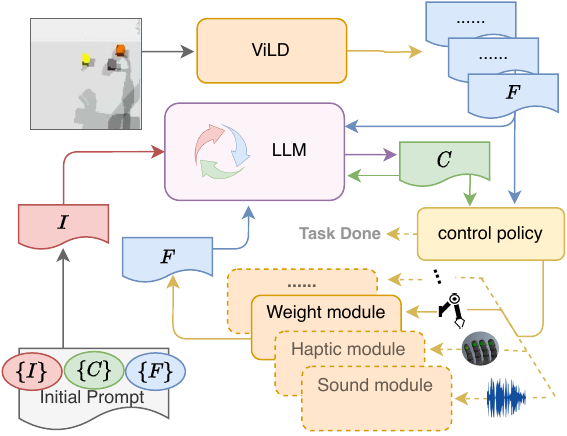}
    \caption{Overview of Matcha. The framework contains an \textcolor{mypurple}{LLM}, \textcolor{myorange}{multimodal perception modules}, and a language-conditioned \textcolor{myyellow}{control policy}. 
    These components communicate with each other with natural language as the intermediate representation. Three types of language information are involved in composing the prompt: 
    \textcolor{myred}{\textit{I}} is a language \textcolor{myred}{instruction} from the user, \textcolor{mygreen}{\textit{C}} is a language \textcolor{mygreen}{command} produced by the LLM, and \textcolor{myblue}{\textit{F}} is semantic \textcolor{myblue}{feedback} from multimodal perceptions. Dotted lines indicate possibly evoking paths.}
    \label{fig:overview-architecture2}
\end{figure}

Fig.~\ref{fig:overview-architecture2} provides an overview of the architecture of Matcha agent. It is a modular framework of three parts: an LLM backbone, multimodal perception modules, and a low-level command execution policy. They connect to each other with language as the intermediate representation for information exchange. 

To be specific, given a high-level instruction, especially the one that Matcha cannot directly perform with the command policy alone, the LLM backbone will reason the situations and select the most contextually admissible perceiving command to gather information.
After the execution of the policy module, the resulting environmental response is processed by a correspondingly evoked multimodal perception module into semantic descriptions, e.g.\ ``clinking sound" by an auditory module after the ``knock on" action. 
Finally, the executed command itself as well as the environmental state description are fed back to the LLM for future planning.

The LLM is used in a few-shot manner without any need for fine-tuning, being independent of other components.
Policy and perception modules can be separately designed and plugged into the framework whenever needed.
Intrinsically linked by natural language, this framework is flexible and can scale and adapt easily to possible robotic upgrades or diverse robotic scenarios.
\subsection{Multimodal Perception and Execution Policy} 
We select a commonly feasible suit of modalities and a language-conditioned policy as an example implementation of our framework.
Other varieties for specific scenarios can also be easily integrated due to the flexibility of modularity of the framework. Detailed experimental implementations will be introduced in Sec. \ref{sec:experiments}.

\subsubsection{Vision} 
Usually, a monitoring camera is the cheapest option for a robot to passively perceive such rich information.
We employ pre-trained ViLD \cite{Gu2022OpenvocabularyObjectDetection}, an open-vocabulary visual detection model, as the vision perception module to detect objects with their categories and positions in the scene.
Then, the results will be delivered to a policy module for identification and execution.
Meanwhile, a prompt template ``The scene contains [\textit{OBJ1}, \textit{OBJ2}, ...]" is applied to construct a scene description, which enables the LLM to have an initial impression of the environment.
Typically, pre-trained vision models are not designed to discern attributes that extend beyond those easily extractable from topology or textures, such as material composition. 
The use of low-resolution images for expedited processing exacerbates the loss of information concerning such attributes.
In our experimental approach, we prioritize demonstrating the integration of diverse modalities instead of extensively fine-tuning ViLD to encompass all aspects.

\subsubsection{Impact Sound}
Impact sound commonly occurs from time to time, and can be useful for robotic multimodal learning \cite{Zhao22ImpactMakes}. 
Though it can be passively collected with a microphone attached to the robotic end-effector, without intentional intervention by the robot, a ``knock on" action in our case, a microphone may only be able to collect background noise.
This auditory perception module classifies the consequent impact sound into a description and then wraps it in a natural language form. 
Actually, a clip of audio may contain sufficient information for some of the usage, e.g.\ to distinguish metal from glass \cite{Dimiccoli22RecognizingObject}. However, it may not be the case for other scenarios, for example, to select the only targeted one among a set of similar ``dull" sounds that could indicate either plastic, wood or hard paper. Therefore, we showcase both of the designs, i.e.\ one with a specific material classification (e.g.\ ``glass") and another with solely low-level and non-distinct descriptions (e.g.\ ``tinkling").
The modular output is also wrapped with templates to a full sentence such as ``It sounds tinkling", to guarantee processing consistency with LLMs. 

\subsubsection{Weight}
Weight measurements are usually obtained via the torque exerted on the robotic arm subsequent to the execution of an ``weighing" action.
The weight information is directly translated into natural language like ``It is lightweight" or ``It weighs 30g". Note that with implicit clarification of the scenario and the type of objects that a robot is manipulating, LLMs can interpret numerical values into contextual meanings.

\subsubsection{Haptics}
Haptic perception is extremely important for humans to interact with their surroundings. It also provides a potential for robots when acquiring information related to physical properties, including hardness, texture, and so on. However, a high-resolution tactile sensor is costly and not worthwhile for many applications. Therefore, in our case, we only use highly abstract descriptions for the force-torque feedback subsequent to a ``touching" action on an object, e.g.\ ``It feels soft" or ``It feels hard and smooth".

\subsubsection{Execution Policy}
The execution policy is conditioned on the generated command by an LLM and the visual information provided by the vision perception module. 
Once an actionable command together with an identified target is suggested by the LLM, the policy module locates the targeted object and executes a certain action. Meanwhile, the environmental feedback will be concurrently collected for multimodal perception modules for further post-processing as demonstrated above.

\subsection{Prompt Engineering}
An issue of grounding LLMs on robotic scenarios is that some of the suggestions generated by LLMs are not executable for a specific robot \cite{Ahn22CanNot, Huang22LanguageModels}, which stems from the fact that LLMs are pre-trained with extremely large open-domain corpora, while the robot is constrained by its physical capability and application scenarios, e.g. a tabletop robot is not able to perform ``walk" action.

In this work, the LLM is applied for few-shot planning \cite{Mialon23AugmentedLanguage, Zeng23SocraticModels}, in which all the executable commands are defined together with several task examples as the initial ``chat" history. See Tab.~\ref{tab:prompt} for the leading prompt which enables the LLM to ground on the specific scenario and follow the contextual patterns for commanding the execution policy.
\input{init_prompt}

We found that only language models that are large enough can follow the patterns in the prompt strictly, i.e.\ only generate commands that have been defined in strictly case-sensitive letters and with the same amount of allowed parameters for each, while small ones can hardly obey this constraint and generate unexpected commands, which brings extra demands for tuning.
As the action planning is performed by LLMs constrained by a given prompt, 
the proposed framework demonstrates high flexibility and generalizability upon the possible incorporation of novel actions or perception modules into the system.

%% file: init_prompt.tex
\begin{table}[ht!]
\centering
\caption{
The snippet of the 5-shot prompt setting. The other four exemplars are omitted here due to the content limit.}
\begin{adjustwidth}{-0.2cm}{0.2cm}
\begin{tabularx}{\linewidth}{l}
\begin{tcolorbox}[
    fonttitle=\fontsize{9}{12}\selectfont, 
    colbacktitle=gray,
    colframe=gray,
    boxrule=0.5pt,
    standard jigsaw,
    opacityback=0,  
    frame hidden,
    interior hidden,
    boxsep=0pt,
    left=4pt,
    right=4pt,
    top=4pt,
    bottom=4pt,
    fontupper=\linespread{0.8}\selectfont,
    fontlower=\linespread{0.8}\selectfont,
]
{\small
\textcolor{gray}{
The followings are conversations with an AI to complete tasks that require active information gathering from multimodalities. Otherwise, the materials of objects are unknown, and it will be ambiguous for an AI to choose the right object.
AI has the following skills to help complete a task:}\\
\textcolor{gray}{1. ``robot.knock\_on()'': to knock on any object and hear the sound to determine the material it consists of. Most of the materials can be determined by this skill.}\\
\textcolor{gray}{2. ``robot.touch()'': to touch with haptics sensors. It is useful for some of the materials.}\\
\textcolor{gray}{3. ``robot.weigh()'': to weigh objects if the knocking method is not proper.}\\
\textcolor{gray}{4. ``robot.pick\_up()'': to pick up the targeted object. After this skill is performed, the episode will terminate with the result.}\\
\textcolor{gray}{Note that the tasks are always set to be accomplishable, and the selected skill should start with a ``$>$'' symbol.} \\
\\
... \\
\\
\textcolor{gray}{Human:} \textcolor{gray}{``pick up the glass block" in the scene contains [yellow block, blue block, green block]}\\
\textcolor{gray}{AI:} \textcolor{gray}{\textit{robot.weigh(yellow block)}}\\
\textcolor{gray}{Feedback:} \textcolor{gray}{It weighs light.}\\
\textcolor{gray}{AI:} \textcolor{gray}{\textit{robot.weigh(blue block)}}\\
\textcolor{gray}{Feedback:} \textcolor{gray}{It weighs a little bit heavy.}\\
\textcolor{gray}{AI:} \textcolor{gray}{\textit{robot.knock\_on(blue block)}}\\
\textcolor{gray}{Feedback:} \textcolor{gray}{It sounds tinkling.}\\
\textcolor{gray}{AI:} \textcolor{gray}{\textit{robot.pick\_up(blue block)}}\\
\textcolor{gray}{done() \\
...}
}
\end{tcolorbox}
\label{tab:prompt}
\end{tabularx}
\end{adjustwidth}
\end{table}

%% file: experiments.tex
\section{Experiments}
\label{sec:experiments}
\subsection{Experimental Setup}
We evaluate the proposed framework in an object-picking task: a robot is instructed to pick up an object that is referred to by a latent property -- \textit{material} -- which is, however, not visually distinguishable under our settings. 
Tasks are intentionally designed such that information from a single modality could be insufficient to determine object properties, while other perception sources can provide compensations to reduce or eliminate this ambiguity.
For example, glass and metal surfaces could exhibit similar hard and smooth properties upon contact, in which case differences in impact sound can aid in further differentiation.
Tab.~\ref{table:metarials-properties} lists variational multimodal descriptions of the materials. These properties are wrapped as natural language sentences before being fed back to the LLM.

Experiments are done in CoppeliaSim\footnote{\url{https://www.coppeliarobotics.com/}} simulations with the NICOL robot \cite{Kerzel23NICOLNeuroinspired},
where several blocks in various colors, materials, weights, and surface textures are randomly selected and placed on the table next to a brown container (see Fig.~\ref{fig:exp}).
The ViLD \cite{Gu2022OpenvocabularyObjectDetection} model is meant to be easily generalized to describe complex scenes despite the simplicity of the object setting here.
After detection, the objects are represented universally by their name, which serves as a parameter for the action function to identify.
Objects with the same color will be distinguished as ``.. on the left/right'' given the simplicity of avoiding more than two duplicated colors for the same shape.
The desktop robot is equipped with two \textit{Open-Manipulator-Pro} arms \footnote{\url{https://emanual.robotis.com/docs/en/platform/openmanipulator\_p/overview/}}, but only its right arm is activated to operate.
It is capable of executing actions in [``knock on", ``touch", ``weigh", ``pick up"], with a parameter to indicate the targeted object.
The first three actions correspond to the interactive perception of impact sound, haptics, and weight respectively, and the last action finalizes the task by picking and transporting an object into the box.
Each instruction is guaranteed to be achievable with the capability of the robot. 

Due to the lack of support for physics-driven sound and deformable object simulation in Coppeliasim, we have implemented reasonable alternatives.
For the haptics of objects, we simplify haptic perception by assigning variational descriptions regarding its material, e.g.\ fibrous objects are usually perceived as ``soft" and a plastic object can be either ``soft" or ``hard". Note that advanced implementations can also be achieved using a neural network as is used in the sound perception module when haptics data for deformable objects is available. 
For the impact sound, we split the YCB-impact-sound dataset \cite{Dimiccoli22RecognizingObject} into training and testing sets and augment them with tricks such as shifting, random cropping, and adding noise.
The training set is used to train our auditory classification neural networks,
while the audios in the testing part are randomly loaded as an alternative to run-time impact sound simulation for the materials mentioned,

Sound can be informative, though not perfect, for determining materials \cite{Dimiccoli22RecognizingObject}.
Besides showing the mediating ability of multiple modalities by the LLM, we further investigate its reasoning ability by employing indistinct descriptions instead of exact material labels.
\begin{itemize}
    \item  \textit{Distinct description}: the sound module describes sound feedback by the corresponding material name and its certainty from the classification model, e.g. ``It is probably glass" or ``It could be plastic with a 47\% chance, or ceramic with a 35\% chance". 
    The distinct description setting is more task-oriented, and it examines the robot's ability to mediate multiple sensory data for disambiguation.
    \item \textit{Indistinct description}: we listed some commonly used indistinct sound descriptions in human communications in Tab.~\ref{table:metarials-properties}, e.g.\ using ``dull" to describe the sound from a plastic block and ``tinkling" to describe the sound for both ceramic and glass objects.
    This setting is more task-agnostic and thus has the potential for generalization. 
    Moreover, it compels the LLM to infer ``professional" material terminology from ambiguous yet multimodal descriptions. 
\end{itemize}
The online OpenAI \textit{text-davinci-003} API\footnote{\label{footnote:gpt3-models} \url{https://platform.openai.com/docs/models/gpt-3}} is applied as the LLM backend because it demonstrates robust instruction-following ability and outstanding reasoning performance.\footnote{The \textit{code-davinci-002} is not chosen because it is common sense instead of the ability of code completion that matters to the active perception. Upon the time that this experiment was carried out, the \textit{text-davinci-003} model is the state-of-the-art GPT-3.5 model available; 
while the later released ChatGPT or GPT-4 model showcases the impressive improved abilities of reasoning, future works will explore the potential of these models.}
We also evaluate with a weaker but far less expensive LLM \textit{text-ada-001}, a GPT-3 model which is usually fast and capable of simple tasks, under the same setting as comparison.
\begin{table}[ht!]
\begin{tabularx}{\linewidth}{|l|X|X|X|}
\hline
\multicolumn{1}{|c}{\textbf{Materials}} & \multicolumn{1}{|c}{\textbf{Impact Sound}} & \multicolumn{1}{|c}{\textbf{Haptics}} & \multicolumn{1}{|c|}{\textbf{Weight}}  \\ \hline \hline
Metal & ``resonant and echoing", ``metallic", ``ringing" & ``hard and cold", ``rigid, cold, and smooth" & ``heavy", ``300g" \\\hline
Glass  &  ``tinkling", ``tinkling and brittle" & ``hard", ``hard and smooth", ``cold and smooth" & ``a little bit heavy", ``150g" \\ \hline
Ceramic & ``clinking and rattling", ``rattling", ``tinkling and brittle" & ``hard", ``tough" &  ``average weight", ``not too light nor not too heavy", ``100g"  \\ \hline
Plastic  & ``dull", ``muffled" & ``hard", ``soft" &  ``light", ``30g" \\ \hline
Fibre & ``muted", ``silent" & ``soft", ``flexible" & ``lightweight", ``underweight", ``10g" \\ \hline
\end{tabularx}
\caption{\label{table:metarials-properties}  Property descriptions of different materials. }
\end{table}
\subsection{Results}
We test the proposed framework Matcha in 50 randomly generated scenarios for each setting and report the success rate. 

We report that the impact sound classification model pre-trained with the selected materials achieves an accuracy of 93.33\%.
When using distinct descriptions, suppose we are making hard-coded rules to utilize the sound module to identify the targeted material, the robot can randomly knock on an object among three, and classify the material until the one that is classified as the target. In theory, the success rate computes as  $\frac{1}{3} p + \frac{2}{3} p^2|_{p=93.33\%}=89.18\%$, where $p$ is the modular accuracy. Usually, other modalities, in this case, are not as distinct as sound, and it could be non-ideal for humans to craft such fusion rules for a possible slight improvement. Therefore, the theoretical success rate 
with only the sound module will be used as our baseline for analysis. Note that this is a reasonable rule that humans will follow, thus it can also be regarded as the upper bound for Matcha if it worked with only impact sound.

Unsurprisingly, Matcha achieves a relatively higher success rate of 90.57\% compared to the ideal theory baseline, as it utilizes compensatory information from other modalities in addition to sound.
When using the indistinct description of impact sound, Matcha is still able to achieve a success rate of 56.67\%, which is larger than a chance success rate of 33.33\% achieved by randomly picking one from the three.
This result is remarkable as it performs few-shot deduction with only indistinct adjectives available.
By analyzing the failure cases, we found that the similar descriptions of glass and ceramic in terms of impact sound, haptics, and weight make it challenging to distinguish one material from the other.
This is not an issue with distinct descriptions, where the sound classification network directly predicts the material, preventing a significant loss of information in the translation from sensory data to language.
While the system displays remarkable reasoning ability, the results still highlight the significance of a well-designed modular description that is not overly abstract in facilitating interactive perceptions.

We observe that only strong LLMs (GPT-3 \textit{text-davinci-003} in our experiments) can perform the task following the primary prompt as we defined, while weak ones possibly generate commands that are not executable and show degraded performance (see Tab.~\ref{table:rate}). 
For example, GPT-3 \textit{text-ada-001}\textsuperscript{\ref{footnote:gpt3-models}}, a smaller model in the GPT-3 model series, may generate commands like ``\textit{robot.knock\_on(metal block)}" or ``\textit{robot.weigh(yellow block, blue block)}", which is not feasible for the policy of the robot. 
\begin{table}[ht!]
\centering
\begin{tabularx}{0.8\linewidth}{lcc}
\multicolumn{1}{c}{\textbf{LLM}}& \textbf{Type of Description} & \textbf{Success Rate} \\
\hline
\multirow{2}{*}{text-ada-001} & Indistinct & 19.05\% \\
 & Distinct & 28.57\% \\ 
 \hline
\multirow{2}{*}{text-davinci-003} & Indistinct & 56.67\% \\
 & Distinct & 90.57\% \\
\hline
\end{tabularx}
\caption{Effect of different LLMs on success rate. \label{table:rate}}
\end{table}
\subsection{Case Studies}
\begin{figure}[thpb]
    \centering
    \includegraphics[width=0.97\linewidth]{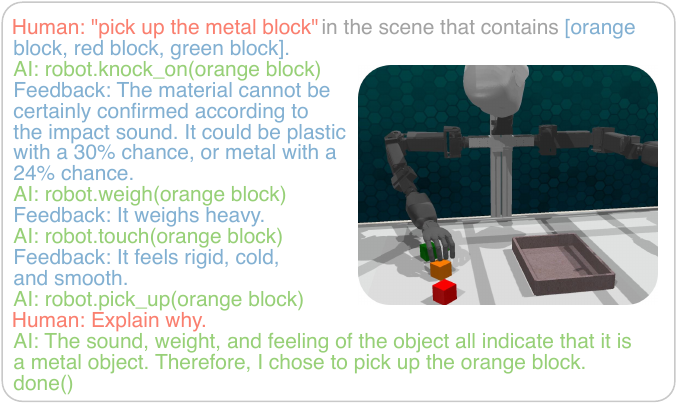}
    \caption{A successful example with a distinct description of impact sound. This example shows that by leveraging multimodal perception, LLM rectifies the misclassification that may occur when relying solely on sound modules.}
    \label{fig:case2}
\end{figure}

\begin{figure}[thpb]
    \centering
    \includegraphics[width=0.97\linewidth]{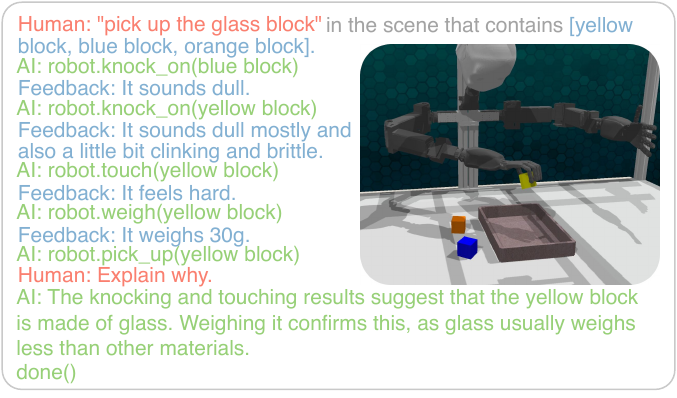}
    \caption{An example in which the agent fails to distinguish glass and ceramic in the setup of using indistinct descriptions of impact sound.}
    \label{fig:case3}
\end{figure}

We provide case studies to showcase the interactive perception ability of Matcha in multimodal environments.
Following the convention of this paper, texts containing information from humans, LLMs, and the environment are indicated in red, green, and blue, respectively.
\begin{figure}[thpb]
    \centering
    \includegraphics[width=0.97\linewidth]{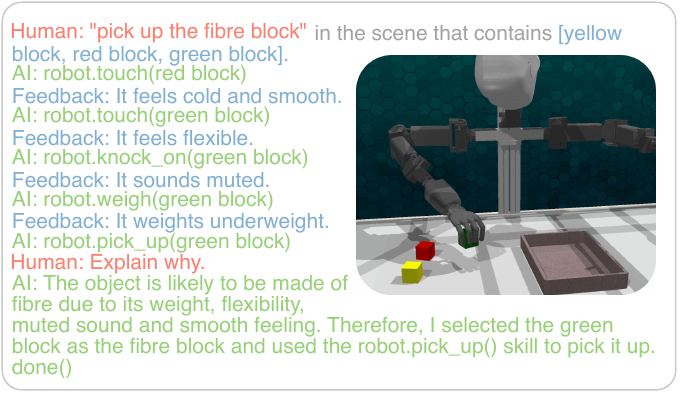}
    \caption{A successful example in which the robot deduces ``fiber" material with indistinct descriptions of impact sound.}
    \label{fig:case1}
\end{figure}
Fig.~\ref{fig:case1} gives a completion that the agent can deduce the latent material property from multimodal yet indistinct descriptions.
During the planning process, the agent decidedly terminates exploration of the ``red block" and instead engages in interacting with the ``green block" exhaustively for a comprehensive examination.
Common sense and, moreover, established knowledge in the LLM enables Matcha to efficiently interact with the environment.
Matcha's proficient behaviors provide evidence of effective reasoning, as it aligns with the subsequent explanation provided by the LLM, namely, that fiber can often be considered ``flexible" rather than ``cold and smooth".
The example depicted in Fig.~\ref{fig:case2} presents a fascinating observation: the impact sound of the ``orange block" suggests it is more likely to be plastic than metal, but Matcha accurately distinguishes it from plastics after engaging in the interactive perception of weight and haptics. This showcases the potential of multimodal perception to improve classification accuracy.
Fig.~\ref{fig:case3} provides a failure case with indistinct descriptions where the robot picks up a ceramic block when asked to pick up the one made of glass.
The underlying cause of this failure is the sensing similarity between glass and ceramic, which creates difficulty in resolving epistemic uncertainty.
\subsection{Discussion}
Weak LLMs, e.g. ones without fine-tuning on instruction alignment \cite{Ouyang22TrainingLanguage}, may not have sufficient capability for precise planning, and thus may require carefully engineered prompts or other grounding techniques.
On the other hand, strong LLMs exhibit impressive in-context learning abilities.
These observations highlight the potential of leveraging knowledge within strong LLMs, as it enables the successful execution of tasks that were previously deemed infeasible.
LLMs can derive significant advantages from utilizing common knowledge, being robust to various instructions regardless of their changes in synonym, linguistic structure or even semantic meanings out of the scope that the robot is initially designed within,
e.g.\ an instruction variation from ``the metal block" to ``a block that may be suitable for cracking a nut",
in which the robot has to establish a meaningful connection between the object's multimodal perceptions and the required utility.

Nevertheless, the reasoning trace may not always align with human expectations.
There are cases that LLMs may prematurely draw conclusions due to their limited logical reasoning ability, particularly when faced with a task that requires reasoning from a long list of facts.

%% file: limitation.tex
\section{Generalization, Limitation and Future Work}
The Matcha framework exhibits a high degree of generalizability thanks to the commonsense knowledge inside LLMs.
Without LLMs, a control algorithm, e.g. one trained with reinforcement learning \cite{Li23InternallyRewarded, Singh20COGConnecting}, may require massive datasets/interactions to learn
the common sense \cite{Singh20COGConnecting} of collaborating different modalities, yet being less efficient and generalizable.

However, interpreting the real world with language can be limited to the complexity of the task and the environment dynamics.
For example, advanced reasoning techniques such as decomposing may be required to deal with a complicated task,
where the task is decomposed into several sub-tasks to tackle separately. 
This automatic operation highlights the flexibility of LLMs but also poses challenges to the static language expression of a complex world
--- The vision-to-language module should be called multiple times with flexible queries.
This brings the requirement of vision-enabled LLMs \cite{Zhu23MiniGPT4Enhancing, Brohan23RT2Visionlanguageaction}, 
built on which the reasoning can be malleable. But multimodal LLMs are yet less controllable and accurate in terms of describing the scene
compared with a templated module.

Despite current limitations, multimodal LLMs gain increasing attention due to their great potential and flexibility.
Future work will explore the multimodal models \cite{Tong22VideoMAEMasked, Brohan23RT2Visionlanguageaction} to leverage unified features.

%% file: conclusion.tex
\section{Conclusions}

LLMs have shown their impressive ability in language generation and human-like reasoning. 
Their potential for integration and enhancement with other fields has attracted growing attention from different research areas.
In this work, we demonstrate the superiority of using an LLM to realize interactive multimodal perception. 
We propose \textbf{Matcha}, a multimodal interactive agent augmented with LLMs, and evaluate it on the task of uncovering object-latent properties. 
Experimental results suggest that our agent can perform interactive multimodal perception reasonably by taking advantage of the commonsense knowledge residing in the LLM, being generalizable due to its modularity and flexibility.

While strong LLMs perform well for tasks that require general knowledge, training and maintaining LLMs locally is currently costly, given the large computation and memory resources required by such models.
Future works will involve distilling the domain-specific knowledge from LLMs into more manageable local models, which can offer greater flexibility and control while maintaining high levels of performance for robotic applications. 
Furthermore, there is a necessity for additional investigation of prompt engineering and multimodal LLMs to augment the ability for complex dynamics in the real world.